\documentclass{article}
\usepackage{arxiv}

\usepackage[
backend=biber,
style=nature,
citestyle=nature
]{biblatex}

\usepackage{lineno}
\usepackage{enumerate}
\usepackage{url}
\usepackage{xcolor}
\usepackage[breaklinks=true,colorlinks,citecolor=citecolor,bookmarks=false,breaklinks,hyperfootnotes=false]{hyperref}
\definecolor{citecolor}{RGB}{42, 130, 90}
\usepackage{amsmath}
\usepackage{amsfonts}
\usepackage{algpseudocode}
\usepackage{booktabs} % for table
\usepackage{multirow}
\usepackage{layouts}
\usepackage[font=small]{caption}
\usepackage{appendix}

\usepackage{subcaption}
\usepackage{graphicx}
\usepackage{xcolor}

\usepackage{afterpage}
\usepackage{cleveref}
\usepackage{xspace}

\usepackage{algorithm}
\usepackage{algorithmicx}
\usepackage{algpseudocode}

\newcommand{\scalefig}{0.85}
\title{Embodied Intelligence via Learning and Evolution}
\author{Agrim Gupta\textsuperscript{1} \And  Silvio Savarese\textsuperscript{1} \And Surya Ganguli\textsuperscript{2,3,4} \And Li Fei-Fei\textsuperscript{1,4}}

\begin{document}
\maketitle
%%%%%%%%%%%%%%%%%%%%%%%%%%%%%%%%%%%%%%%%%%%%%%%%%%%%%%%%%%%%%%%%%%%%%%%%%%%%%%%%%%%%%%%%%%%%%%%%%%%%%%%%%%%%%%%%%%%%%
%%% Abstract
%%%%%%%%%%%%%%%%%%%%%%%%%%%%%%%%%%%%%%%%%%%%%%%%%%%%%%%%%%%%%%%%%%%%%%%%%%%%%%%%%%%%%%%%%%%%%%%%%%%%%%%%%%%%%%%%%%%%%
\begin{abstract}
 The intertwined processes of learning and evolution in complex environmental niches have resulted in a remarkable diversity of morphological forms. Moreover, many aspects of animal intelligence are deeply embodied in these evolved morphologies. However, the principles governing relations between environmental complexity, evolved morphology, and the learnability of intelligent control, remain elusive, partially due to the substantial challenge of performing large-scale {\it in silico} experiments on evolution and learning. We introduce Deep Evolutionary Reinforcement Learning (DERL): a novel computational framework which can evolve diverse agent morphologies to learn challenging locomotion and manipulation tasks in complex environments using only low level egocentric sensory information. Leveraging DERL we demonstrate several relations between environmental complexity, morphological intelligence and the learnability of control.  First, environmental complexity fosters the evolution of morphological intelligence as quantified by the ability of a morphology to facilitate the learning of novel tasks. Second, evolution rapidly selects morphologies that learn faster, thereby enabling behaviors learned late in the lifetime of early ancestors to be expressed early in the lifetime of their descendants. In agents that learn and evolve in complex environments, this result constitutes the first demonstration of a long-conjectured morphological Baldwin effect. Third, our experiments suggest a mechanistic basis for both the Baldwin effect and the emergence of morphological intelligence through the evolution of morphologies that are more physically stable and energy efficient, and can therefore facilitate learning and control.
\end{abstract}

\vspace{5mm}
%%%%%%%%%%%%%%%%%%%%%%%%%%%%%%%%%%%%%%%%%%%%%%%%%%%%%%%%%%%%%%%%%%%%%%%%%%%%%%%%%%%%%%%%%%%%%%%%%%%%%%%%%%%%%%%%%%%%%
%%% Main Text
%%%%%%%%%%%%%%%%%%%%%%%%%%%%%%%%%%%%%%%%%%%%%%%%%%%%%%%%%%%%%%%%%%%%%%%%%%%%%%%%%%%%%%%%%%%%%%%%%%%%%%%%%%%%%%%%%%%%%
Evolution over the last $600$ million years has generated a variety of ``endless forms most beautiful'' \autocite{darwin1859origin} starting from an ancient bilatarian worm \autocite{Evans2020-zn}, and culminating in a set of diverse animal morphologies. Moreover, such animals display remarkable degrees of embodied intelligence by leveraging their evolved morphologies to learn complex tasks. Indeed the field of embodied cognition posits that intelligent behaviors can be rapidly learned by agents whose morphologies are well adapted to their environment \autocite{pfeifer2001understanding,brooks1991new,bongard2014morphology}.  In contrast, the field of artificial intelligence (AI) has focused primarily on disembodied cognition, for example in domains of language \autocite{brown2020language}, vision \autocite{he2016deep} or games \autocite{silver2016mastering}.
\let\thefootnote\relax\footnote{Video available \href{https://youtu.be/MMrIiNavkuY}{here}.}
The creation of artificial embodied agents with well adapted morphologies that can learn control tasks in diverse, complex environments is challenging because of the twin difficulties of: (1) searching through a combinatorially large number of possible morphologies, and (2) the computational time required to evaluate fitness through lifetime learning.  Hence, prior work has either evolved agents in severely limited morphological search spaces\autocite{sims1994evolving, jelisavcic2019lamarckian, auerbach2014environmental, auerbach2014robogen,wang2018neural} or focused on finding optimal parameters given a fixed hand designed morphology\autocite{luck2020data, schaff2019jointly, ha2019reinforcement}. Furthermore, the difficulty of evaluating fitness forced prior work to (1) avoid learning adaptive controllers directly from raw sensory observations \autocite{sims1994evolving, jelisavcic2019lamarckian, auerbach2014environmental, zhao2020robogrammar}; (2) learn hand designed controllers with few $(\leq 100)$ parameters \autocite{sims1994evolving, jelisavcic2019lamarckian, auerbach2014environmental}; (3) learn to predict the fitness of a morphology \autocite{zhao2020robogrammar, wang2018neural}; (4) mimic Lamarckian rather than Darwinian evolution by directly transmitting learned information across generations \autocite{jelisavcic2019lamarckian, wang2018neural}. Moreover, prior works were also primarily limited to the simple task of locomotion over a flat terrain with agents having few degrees of freedom (DoF) \autocite{wang2018neural, auerbach2014environmental} or with body plans composed of cuboids to further simplify the problem of learning a controller\autocite{sims1994evolving, jelisavcic2019lamarckian, auerbach2014environmental}.   

To overcome these substantial limitations, we propose Deep Evolutionary Reinforcement Learning (DERL), (Fig.~\ref{fig:system}a) a computational framework enabling us to simultaneously scale the creation of embodied agents across $3$ axes of complexity: environmental, morphological, and control. DERL opens the door to performing large-scale {\it in silico} experiments to yield scientific insights into how learning and evolution cooperatively create sophisticated relationships between environmental complexity, morphological intelligence, and the learnability of control tasks.  Moreover, DERL also alleviates sample inefficiency of reinforcement learning by creating embodied agents that can not only learn with less data, but also generalize to solve multiple novel tasks. DERL operates by mimicking the intertwined processes of Darwinian evolution over generations to search over morphologies, and neural learning within a lifetime, to evaluate how fast and well a given morphology can solve complex tasks through intelligent control. Our key contributions are: (1) an efficient asynchronous method for parallelizing computations underlying learning and evolution across many computing elements, thereby allowing us to leverage the scaling of computation and models that has been so successful in other fields of AI \autocite{kaplan2020scaling, henighan2020scaling, brown2020language, chen2020big} and bring it bear on the field of evolutionary robotics\autocite{lipson2000automatic}; (2) the introduction of a UNIMAL, a UNIversal aniMAL morphological design space that is both highly expressive yet also enriched for useful controllable morphologies; (3) a paradigm to evaluate the intelligence embedded directly in a morphology by assessing how well each morphology facilitates the speed of reinforcement learning in a suite of test tasks involving tests of stability, agility and manipulation.  We next describe the engineering elements underlying DERL, the scientific insights it empowers, and the capacity for creating more sample efficient and multi-task RL agents it makes possible.  

\section*{DERL: Computational framework for creating embodied agents}
%%%%%%%%%%%%%%%%%%%%%%%%%%%%%%%%%%%%%%%%%%%%%%%%%%%%%%%%%%%%%%%%%%%%%%%%%%%%%%%%%%%%%%%%%%%%%%%%%%%%%%%%%%%%%%%%%%%%%
\begin{figure*}[t]
\centering
\includegraphics[width=\scalefig\linewidth]{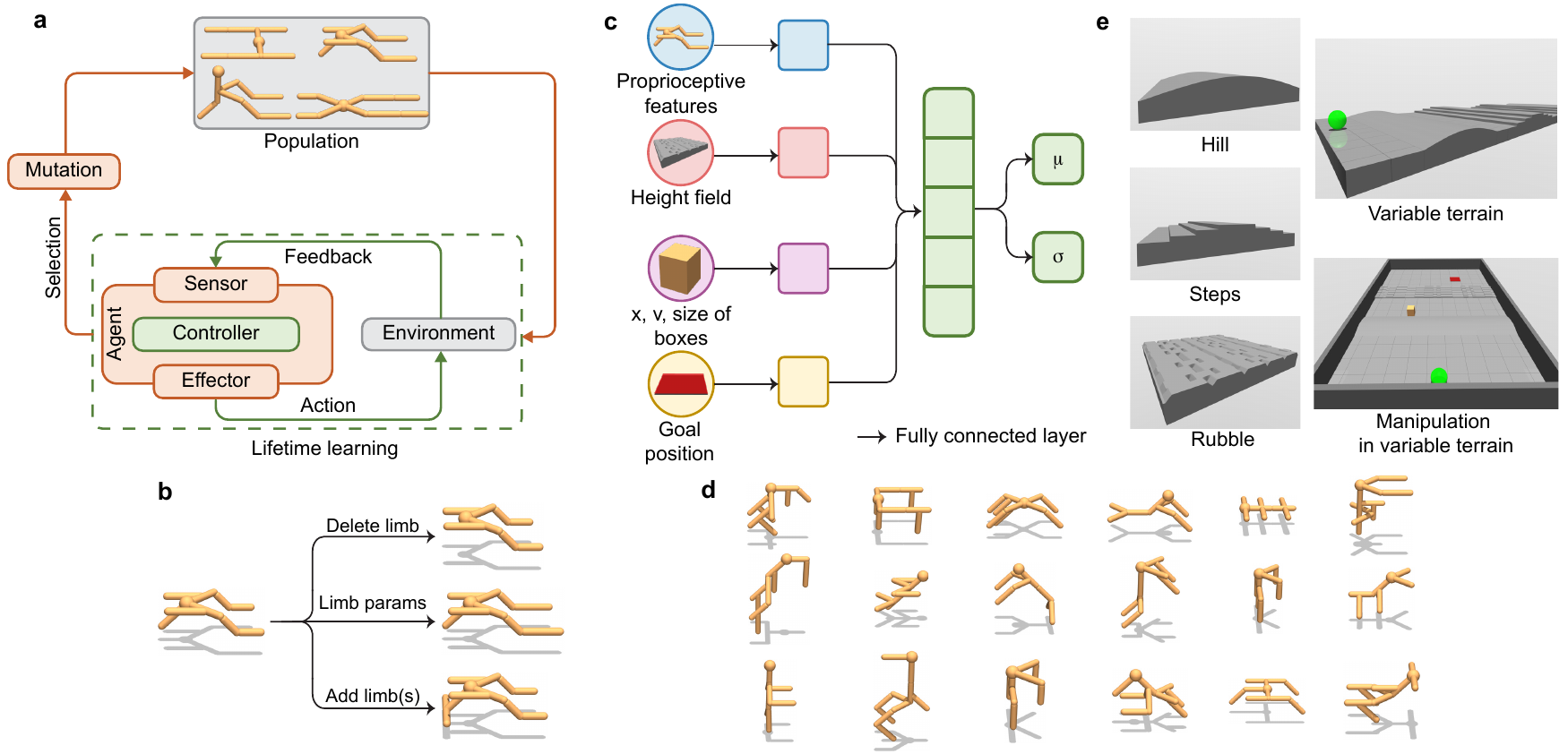}
  \caption{\textbf{DERL overview.}  DERL (\textbf{a}) is a general framework to make embodied agents via two interacting adaptive processes. An outer loop of evolution optimizes agent morphology via mutation operations, some of which are shown in (\textbf{b}) and an inner reinforcement learning loop optimizes the parameters of a neural controller (\textbf{c}). \textbf{d,} Example agent morphologies in the UNIMAL design space. \textbf{e,} Variable terrain consists of three stochastically generated obstacles: hills, steps and rubble. In manipulation in variable terrain, an agent must start from an initial location (green sphere) and move a box to a goal location (red square).}
\label{fig:system}
\vspace{-3mm}
\end{figure*}
%%%%%%%%%%%%%%%%%%%%%%%%%%%%%%%%%%%%%%%%%%%%%%%%%%%%%%%%%%%%%%%%%%%%%%%%%%%%%%%%%%%%%%%%%%%%%%%%%%%%%%%%%%%%%%%%%%%%%
Previous evolutionary simulations \autocite{wang2018neural, zhao2020robogrammar, sims1994evolving, jelisavcic2019lamarckian, auerbach2014environmental} commonly employed generational evolution\autocite{parallelbook}, where in each generation, the entire population is simultaneously replaced by applying mutations to the fittest individuals. However, this paradigm scales poorly in creating embodied agents due to the significant computational burden imposed by training every member of a large population before any further evolution can occur. Inspired by recent progress in neural architecture search \autocite{real2017large, real2019regularized, zoph2016neural}, we decouple the events of learning and evolution in a distributed asynchronous manner using tournament based evolution \autocite{goldberg1991comparative, real2019regularized}.  Specifically, each evolutionary run starts with a population of $P=576$ agents with unique topologies to encourage diverse solutions. The initial population undergoes lifetime learning via reinforcement learning \autocite{sutton2018reinforcement} (RL) in parallel and the average final reward determines fitness. After initialization, each worker (CPUs) operates independently by conducting tournaments in groups of $4$ wherein the fittest individual is selected as a parent, and a mutated copy (child) is added to the population after evaluating its fitness through lifetime learning. To keep the size of the population fixed we consider only the most recent $P$ agents as alive\autocite{real2019regularized}. By moving from generational to asynchronous parallel evolution, we do not require learning to finish across the {\it entire} population before any further evolution occurs. Instead as soon as any agent finishes learning, the worker can immediately perform another step of selection, mutation and learning in a new tournament. 

For learning, each agent senses the world by receiving only low level egocentric proprioceptive and exteroceptive observations and chooses its actions via a stochastic policy determined by the parameters of a deep neural network (Fig.~\ref{fig:system}b) that are learned via proximal policy optimization (PPO) \autocite{schulman2017proximal}. See Appendix B for details about the sensory inputs, neural controller architectures and learning algorithms employed by our agents.

Overall, DERL enables us to perform large-scale experiments across $1152$ CPUs involving on average $10$ generations of evolution that search over and train $4000$ morphologies, with $5$ million agent-environment interactions (i.e. learning iterations) for each morphology. At any given instant of time, since we can train 288 morphologies in parallel asynchronous tournaments, this entire process of learning and evolution completes in less than $16$ hours. To our knowledge this constitutes the largest scale simulations of simultaneous morphological evolution and RL to date.

\section*{UNIMAL: A UNIversal aniMAL morphological design space}
%%%%%%%%%%%%%%%%%%%%%%%%%%%%%%%%%%%%%%%%%%%%%%%%%%%%%%%%%%%%%%%%%%%%%%%%%%%%%%%%%%%%%%%%%%%%%%%%%%%%%%%%%%%%%%%%%%%%%
\begin{figure*}[t]
\centering
\includegraphics[width=\scalefig\linewidth]{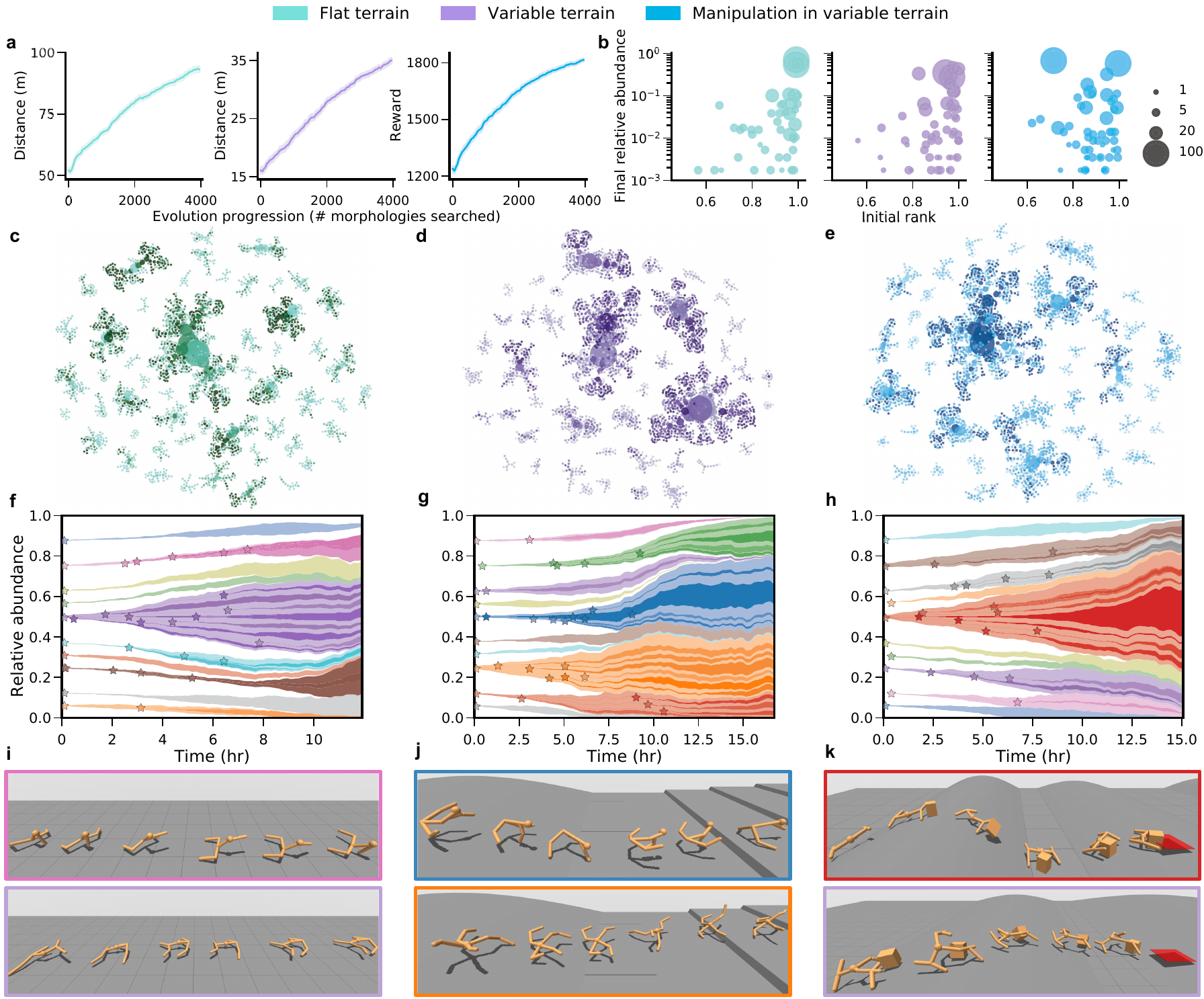}
  \caption{\textbf{Evolutionary dynamics in multiple environments.} \textbf{a,} Mean ($n=300$) and $95\%$ bootstrapped confidence intervals  of the fitness of the top $100$ agents in each of $3$ evolutionary runs. \textbf{b,} Each dot represents a lineage that survived to the end of one of $3$ evolutionary runs. Dot size reflects the total number of beneficial mutations (see Appendix D) accrued by the lineage. The founder of a lineage need not have extremely high initial fitness rank in order for it's lineage to comprise a reasonably high fraction of the final population. It can instead achieve population abundance by accruing many beneficial mutations starting from a lower rank (i.e. large dots that are high and to the left). (\textbf{c-e}) Phylogenetic trees of a single evolutionary run where each dot represents a single UNIMAL, dot size reflects number of descendants, and dot opacity reflects fitness, with darker dots indicating higher fitness.  These trees demonstrate that multiple lineages with descendants of high fitness can originate from founders with lower fitness (i.e. larger lighter dots). (\textbf{f-h}) Muller diagrams \autocite{muller1932some} showing relative population abundance over time (in the same evolutionary run as in \textbf{c-e}) of the top $10$ lineages with the highest final population abundance. Each color denotes a different lineage and the opacity denotes its fitness. Stars denote successful mutations which changed agent topology (i.e. adding/deleting limbs) and resulted in a sub-lineage with more than $20$ descendants. The abundance of the rest of the lineages is reflected by white space. (\textbf{i-k}) Time-lapse images of agent policies in each of the three environments with boundary color corresponding to the lineages above.}
\label{fig:muller}
\vspace{-3mm}
\end{figure*}
%%%%%%%%%%%%%%%%%%%%%%%%%%%%%%%%%%%%%%%%%%%%%%%%%%%%%%%%%%%%%%%%%%%%%%%%%%%%%%%%%%%%%%%%%%%%%%%%%%%%%%%%%%%%%%%%%%%%%
To overcome the limited expressiveness of previous morphological search spaces, we introduce a UNIversal aniMAL (UNIMAL) design space (Fig.~\ref{fig:system}e). Our genotype is a kinematic tree corresponding to a hierarchy of articulated 3D rigid parts connected via motor actuated hinge joints. Nodes of the kinematic tree consist of two component types: a sphere representing the head which forms the root of the tree, and cylinders representing the limbs of the agent. Evolution proceeds through asexual reproduction via three classes of mutation operations (see Appendix A) that: (1) either shrink or grow the kinematic tree by growing or deleting limbs (Fig.~\ref{fig:system}d); (2) modify the physical properties of existing limbs, like their lengths and densities (Fig.~\ref{fig:system}d); (3) modify the properties of joints between limbs, including degrees of freedom (DoF), angular limits of rotation, and gear ratios. Importantly we only allow paired mutations that preserve bilateral symmetry, an evolutionarily ancient conserved feature of all animal body plans originating about $600$ million years ago \autocite{Evans2020-zn}. A key physical consequence is that the center of mass of every agent lies on the saggital plane, thereby reducing the degree of control required to learn left-right balancing.  Despite this constraint, our morphological design space is highly expressive, containing approximately $10^{18}$ unique agent morphologies with less than $10$ limbs.  

\section*{Successful evolution of diverse morphologies in complex environments}
DERL enables us for the first time to move beyond locomotion in flat terrain to simultaneously evolve morphologies and learn controllers for agents in $3$ environments (Fig.~\ref{fig:system}c) of increasing complexity: (1) Flat terrain (FT); (2) Variable terrain (VT); and (3) Non prehensile manipulation in variable terrain (MVT). 
VT is an extremely challenging environment as during each episode a new terrain is generated by randomly sampling a sequence of obstacles. 
Indeed, prior work \autocite{heess2017emergence} on learning locomotion in a variable terrain for a simple $9$ DoF planar 2D walker required $10^7$ agent-environment interactions, despite using curriculum learning and a morphology specific reward function. MVT posses additional challenges since the agent must rely on complex contact dynamics to manipulate the box from a random location to a target location while also traversing VT. See Appendix A for a detailed description of these complex stochastic environments. 

\begin{figure*}[t]
\centering
\includegraphics{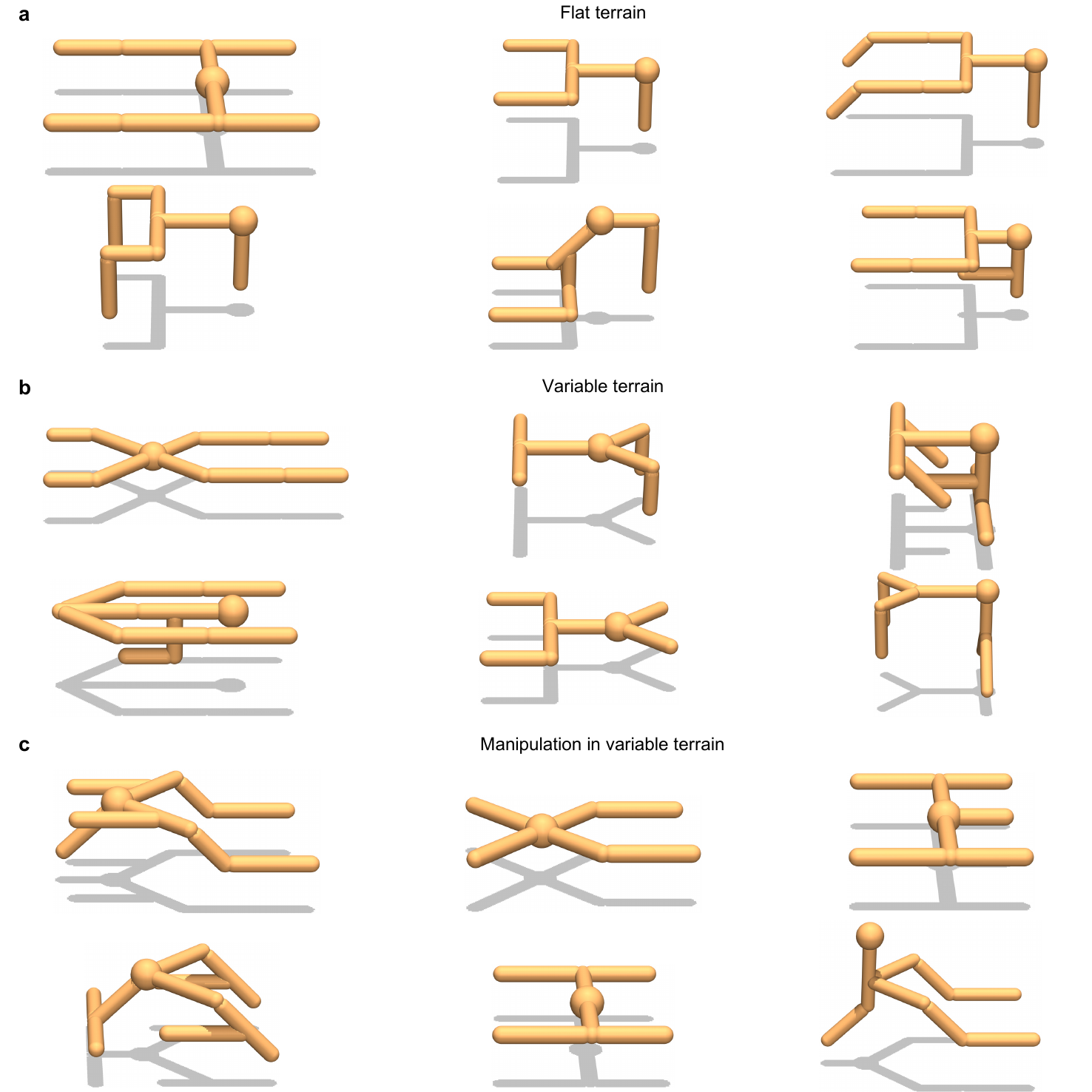}
  \caption{\textbf{Best agent morphologies evolved in different environments.} A subset of the top $10$ agent morphologies evolved across $3$ evolutionary runs. See reporting methodology in Appendix D for details about selection procedure and video for illustration of learnt behaviour.}
\label{fig:top_unimals}
\vspace{-3mm}
\end{figure*}

\begin{figure*}[t]
\centering
\includegraphics{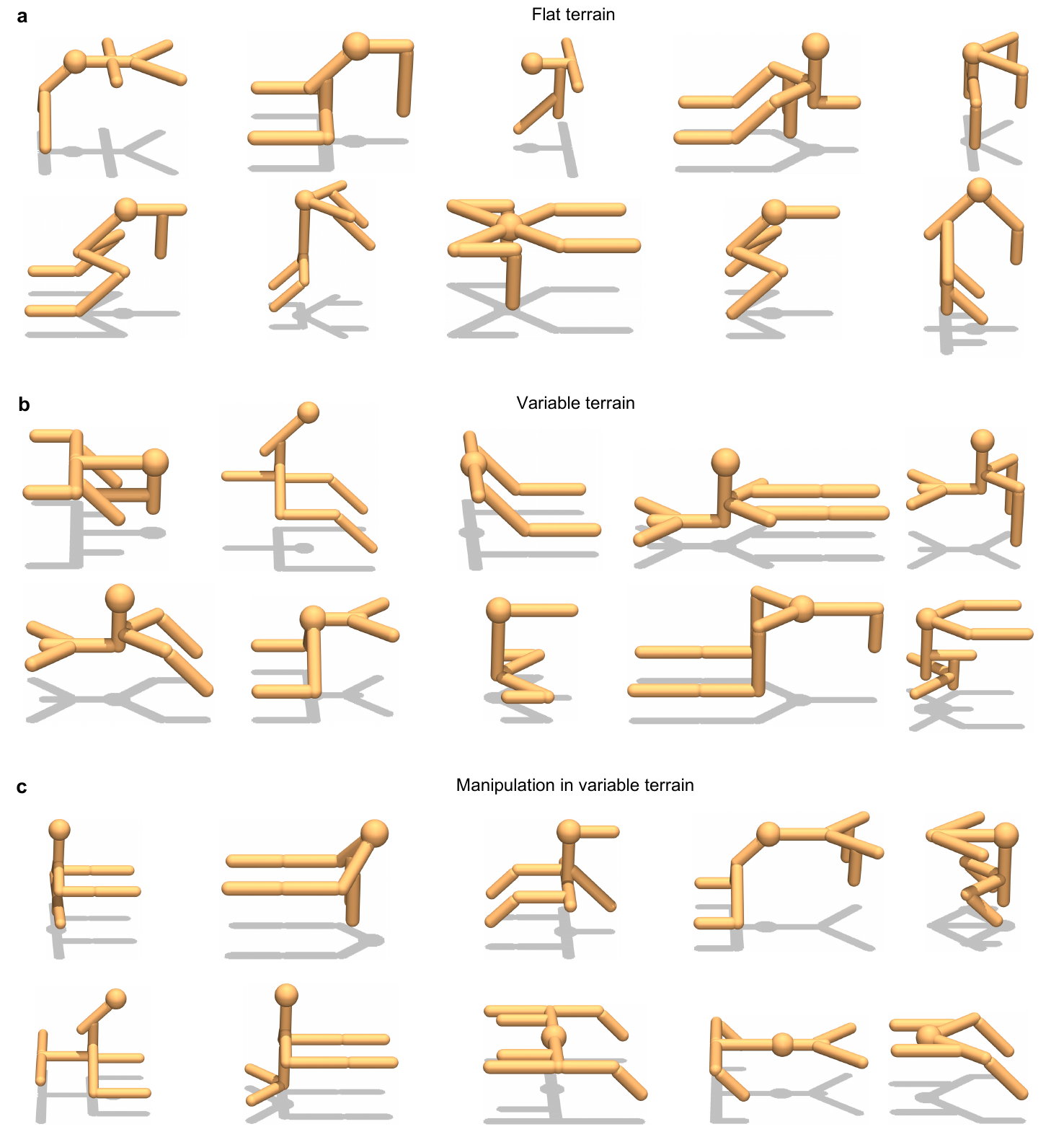}
  \caption{\textbf{Example agent morphologies evolved in different environments.} A subset of the top $100$ agent morphologies evolved across $3$ evolutionary runs. See reporting methodology in Appendix D for details about selection procedure.}
\label{fig:unimal_examples}
\vspace{-3mm}
\end{figure*}

DERL is able to find successful morphological solutions for all $3$ environments (Fig.~\ref{fig:muller}a; see video for illustration of learnt behaviour). Indeed the relatively high average initial fitness before evolution even occurs (Fig.~\ref{fig:muller}a) reflects the efficacy of the UNIMAL design space.  Moreover DERL finds a diversity of successful solutions (Fig.~\ref{fig:muller}b-h). Maintaining solution diversity is generically challenging for most evolutionary dynamics, as often only $1$ solution and its nearby variations dominate. In contrast, by moving away from generational evolution in which the entire population competes simultaneously to survive in the next generation, to asynchronous parallel small tournament based competitions, DERL enables ancestors with lower initial fitness to still contribute a relative large abundance of highly fit descendants to the final population (Fig.~\ref{fig:muller}b). Given the initialized population exhibits morphological diversity, this evolutionary dynamics, as visualized by both phylogenetic trees (Fig.~\ref{fig:muller}c-e) and Muller plots \autocite{muller1932some} (Fig.~\ref{fig:muller}f-h), thereby ensures final population diversity without sacrificing fitness. Indeed, the set of evolved morphologies include different variations of bipeds, tripeds, and quadrupeds with and without arms  (Fig.~\ref{fig:muller}i-k, \ref{fig:top_unimals}, \ref{fig:unimal_examples}).
%%%%%%%%%%%%%%%%%%%%%%%%%%%%%%%%%%%%%%%%%%%%%%%%%%%%%%%%%%%%%%%%%%%%%%%%%%%%%%%%%%%%%%%%%%%%%%%%%%%%%%%%%%%%%%%%%%%%%
\begin{figure*}[t]
\centering
\includegraphics[width=\scalefig\linewidth]{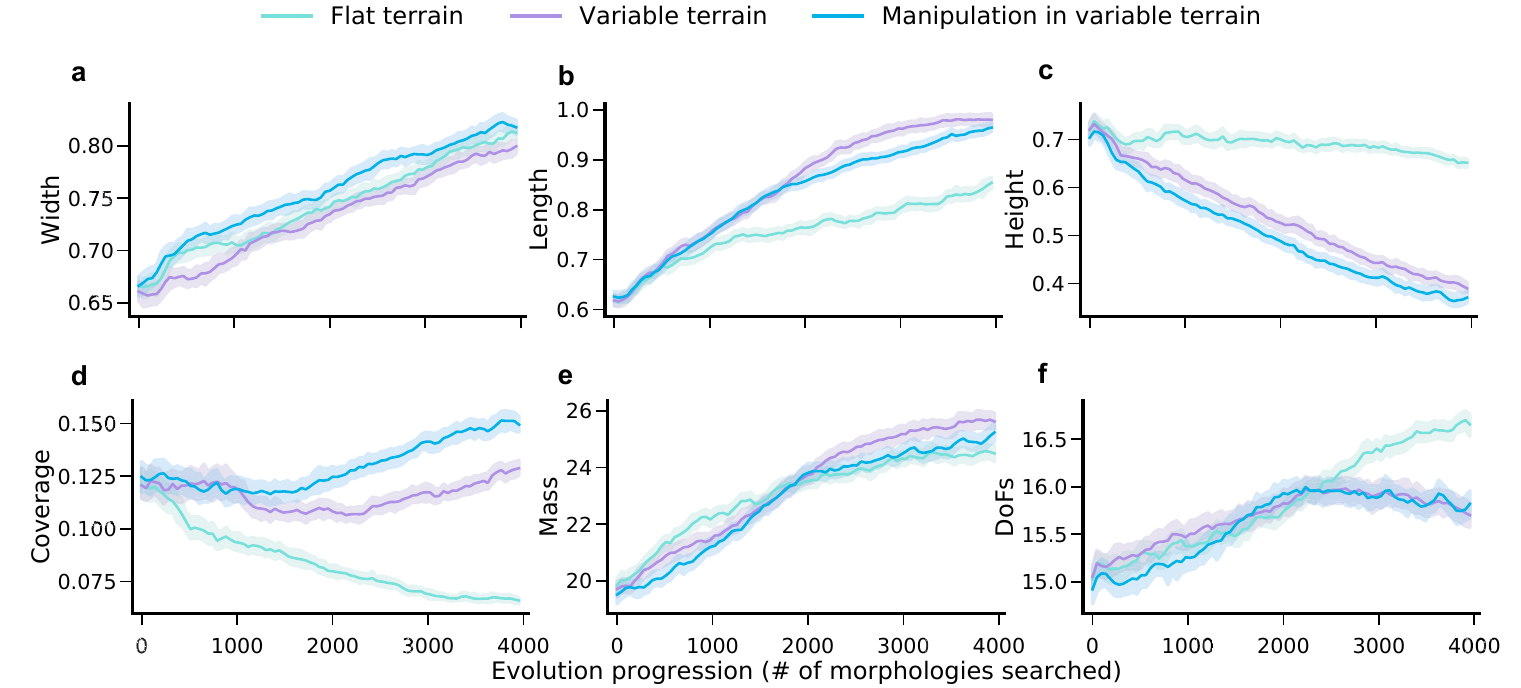}
  \caption{\textbf{Influence of environment on different morphological descriptors.} (\textbf{a-e}) Progression of the mean of different morphological descriptors averaged over $3$ runs in different environments for the entire population. Shaded region denotes $95\%$ bootstrapped confidence interval. (\textbf{a-c}) VT/MVT agents tend to be longer along the direction of forward motion and shorter in height compared to agents evolved in FT. \textbf{d,} Coverage is the ratio of volume of the agent morphology and it's axis aligned bounding box. FT agents are less space filling as compared to VT and MVT agents. Agents evolved in all three environments have the similar masses (\textbf{e}) and DoFs (\textbf{f}).}
\label{fig:morphological_descriptior}
\vspace{-3mm}
\end{figure*}
%%%%%%%%%%%%%%%%%%%%%%%%%%%%%%%%%%%%%%%%%%%%%%%%%%%%%%%%%%%%%%%%%%%%%%%%%%%%%%%%%%%%%%%%%%%%%%%%%%%%%%%%%%%%%%%%%%%%%

We analyze the progression of different morphological descriptors across the $3$ environments (Fig.~\ref{fig:morphological_descriptior}), finding a strong impact of environment on evolved morphologies. While agents evolved in all environments have similar masses and control complexity (as measured by DoF $\approx 16$), VT/MVT agents tend to be longer along the direction of forward motion and shorter in height compared FT agents. FT agents are less space filling compared to VT/MVT agents, as measured by the coverage\autocite{auerbach2012relationship, miras2020environmental} of the morphology. The less space-filling nature of FT agents reflects a common strategy to have limbs spaced far apart on the body giving them full range of motion (Fig.~\ref{fig:muller}i, \ref{fig:top_unimals}a, \ref{fig:unimal_examples}a). Agents in FT exhibit both a falling forward locomotion gait and a lizard like gait (Fig.~\ref{fig:muller}i). Agents evolved in VT are often similar to FT but with additional mechanisms to make the gait more stable. For example, instead of having a single limb attached to the head which breaks falls and propels the agent forward, VT agents have two symmetrical limbs providing greater stability and maneuverability (Fig.~\ref{fig:muller}j,k,\ref{fig:top_unimals}a, b). Finally, agents in MVT develop forward reaching arms mimicking pincer or claw like mechanisms that enable guiding a box to a goal position (Fig.~\ref{fig:muller}k, \ref{fig:top_unimals}c, \ref{fig:unimal_examples}c).

\section*{Environmental complexity engenders morphological intelligence}
%%%%%%%%%%%%%%%%%%%%%%%%%%%%%%%%%%%%%%%%%%%%%%%%%%%%%%%%%%%%%%%%%%%%%%%%%%%%%%%%%%%%%%%%%%%%%%%%%%%%%%%%%%%%%%%%%%%%%
\begin{figure*}[t]
\centering
\includegraphics[width=\scalefig\linewidth]{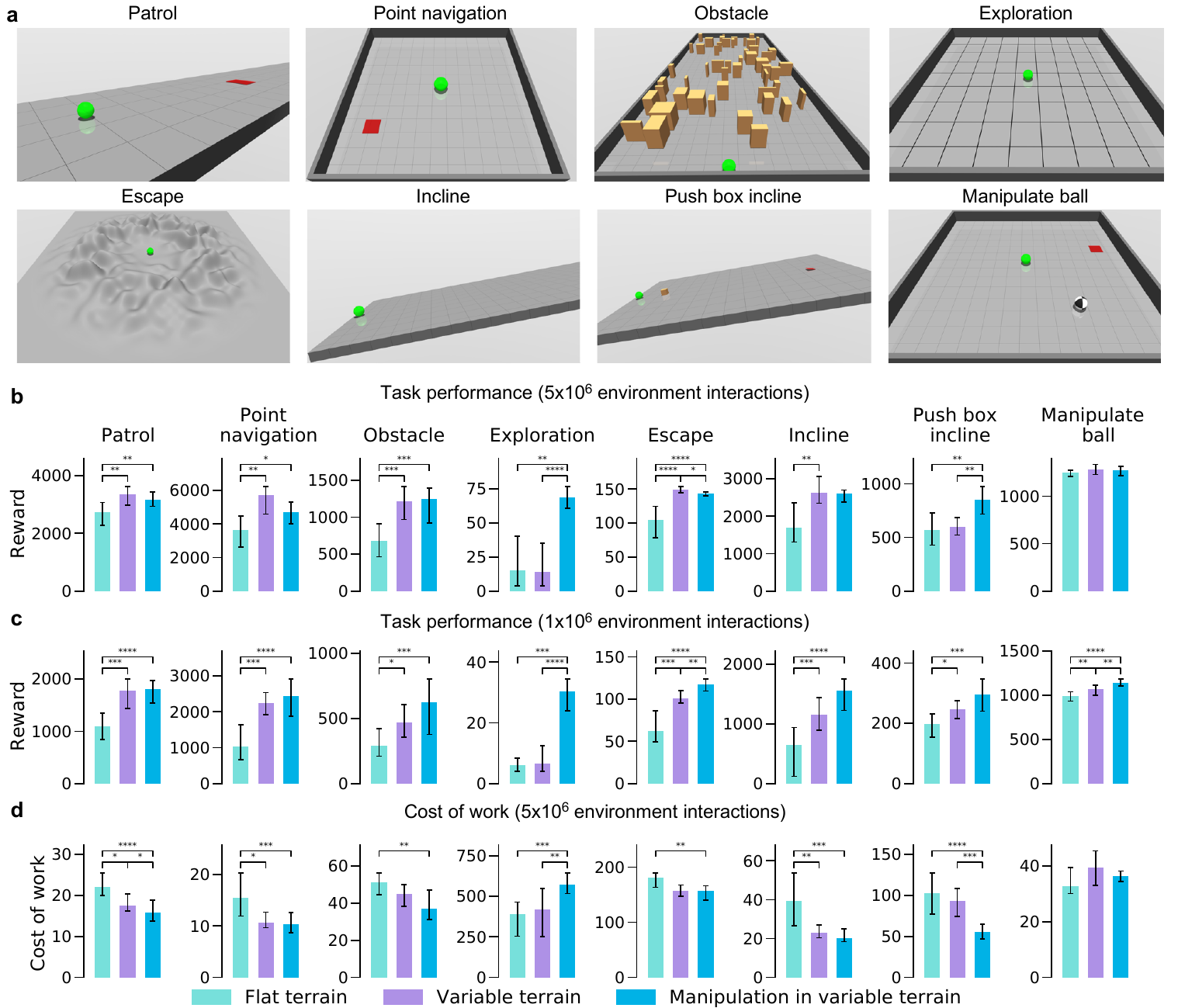}
  \caption{\textbf{Environmental complexity fosters morphological intelligence.} \textbf{a,} Eight test tasks for evaluating morphological intelligence across $3$ domains spanning stability, agility and manipulation ability. Initial agent location is specified by a green sphere, and goal location by a red square (see Appendix C for detailed task descriptions). (\textbf{b-d}) We pick the $10$ best performing morphologies across $3$ evolutionary runs per environment. Each morphology is then trained from scratch for all $8$ test tasks with $5$ different random seeds. Bars indicate median reward ($n=50$) (\textbf{b-c}) and cost of work (\textbf{d}) with error bars denoting $95\%$ bootstrapped confidence intervals and color denoting evolutionary environment.  \textbf{b,} Across $7$ test tasks, agents evolved in MVT perform better than agents evolved in FT. \textbf{c,} With reduced learning iterations ($5$ million in (\textbf{b}) vs $1$ million in (\textbf{c})) MVT/VT agents perform significantly better across all tasks. \textbf{d,} Agents evolved in MVT are more energy efficient as measured by lower cost of work despite no explicit evolutionary selection pressure favoring energy efficiency. Statistical significance was assessed using the two-tailed Mann-Whitney U Test; $*P < 0.05$; $**P < 0.01$; $***P < 0.001$; $****P < 0.0001$.}
\label{fig:eval_suite}
\vspace{-3mm}
\end{figure*}
%%%%%%%%%%%%%%%%%%%%%%%%%%%%%%%%%%%%%%%%%%%%%%%%%%%%%%%%%%%%%%%%%%%%%%%%%%%%%%%%%%%%%%%%%%%%%%%%%%%%%%%%%%%%%%%%%%%%%
The few prior analyses of the impact of environment on evolved morphologies have focused on measuring various morphological descriptors \autocite{miras2020environmental} or on morphological complexity \autocite{auerbach2012relationship}. However, a key challenge to designing any intelligent agent lies in ensuring that it can rapidly adapt to any new task. We thus focus instead on understanding how this capacity might arise through combined learning and evolution by characterizing the intelligence embodied in a morphology as a consequence of it's evolutionary environment. Concretely, we compute how much a morphology facilitates the process of learning a large set of test tasks. This approach is similar to evaluating the quality of latent neural representations by computing their performance on downstream tasks via transfer learning \autocite{pratt1991direct, pmlrv119chen20j, he2020momentum}. Thus in our framework, intelligent morphologies by definition facilitate faster and better learning in downstream tasks. We create a suite of $8$ tasks (Fig.~\ref{fig:eval_suite}a; see video for illustration of learnt behaviour) categorized into $3$ domains testing agility (patrol, point navigation, obstacle and exploration), stability (escape and incline) and manipulation (push box incline and manipulate ball) abilities of the agent morphologies. Controllers for each task are learned from scratch, thus ensuring that differences in performance are solely due to differences in morphologies.

We first test the hypothesis that evolution in more complex environments generates more intelligent morphologies that perform better in our suite of test tasks (Fig.~\ref{fig:eval_suite}b). We find that across $7$ test tasks, agents evolved in MVT perform better than agents evolved in FT. VT agents perform better than FT agents in $5$ out of $6$ tasks in the domains of agility and stability, but have similar performance in the manipulation tasks. To test the speed of learning, we repeat the same experiment with $1/5\textsuperscript{th}$ the learning iterations (Fig.~\ref{fig:eval_suite}c). The differences between MVT/VT agents and FT agents are now more pronounced across all tasks. These results suggest that morphologies evolved in more complex environments are more intelligent in the sense that they facilitate learning many new tasks both better and faster.

\section*{Demonstration of a stronger form of the conjectured morphological Baldwin effect}
%%%%%%%%%%%%%%%%%%%%%%%%%%%%%%%%%%%%%%%%%%%%%%%%%%%%%%%%%%%%%%%%%%%%%%%%%%%%%%%%%%%%%%%%%%%%%%%%%%%%%%%%%%%%%%%%%%%%%
\begin{figure*}[t]
\centering
\includegraphics[width=\scalefig\linewidth]{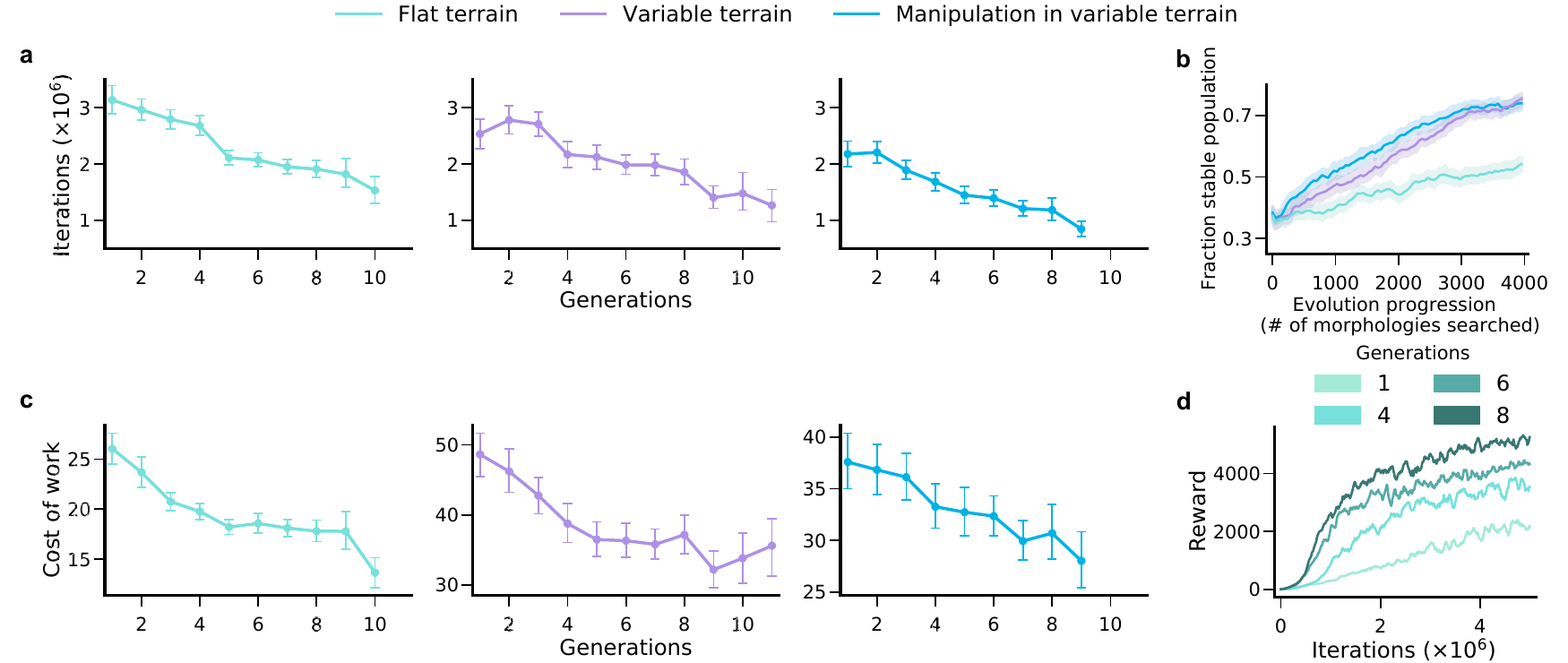}
  \caption{\textbf{A morphological Baldwin effect and its relationship to energy efficiency and stability.} \textbf{a,} Mean ($n=100$) iterations to achieve the $75\textsuperscript{th}$ percentile fitness of the initial population for the top $100$ agents across $3$ evolutionary runs, as a function of generations. \textbf{b,}  Fraction of stable morphologies (see Appendix D) averaged over $3$ evolutionary runs per environment.  This fraction is higher in VT and MVT than FT, indicating that these more complex environments yield an added selection pressure for stability.  \textbf{c,} Mean cost of work (see Appendix D) for same top $100$ agents as in \textbf{a}. \textbf{d,} Learning curves for different generations of an illustrative agent evolved in FT indicate that later generations not only perform better but also learn faster. Thus overall evolution simultaneously discovers morphologies that are more energy efficient (\textbf{c}), stable (\textbf{b}) and simplify control, leading to faster learning (\textbf{a}). Error bars (\textbf{a, c}) and shaded region (\textbf{b}) denote $95\%$ bootstrapped confidence interval.}
\label{fig:baldwin}
\vspace{-3mm}
\end{figure*}
%%%%%%%%%%%%%%%%%%%%%%%%%%%%%%%%%%%%%%%%%%%%%%%%%%%%%%%%%%%%%%%%%%%%%%%%%%%%%%%%%%%%%%%%%%%%%%%%%%%%%%%%%%%%%%%%%%%%%
The coupled dynamics of evolution over generations and learning within a lifetime have long been conjectured to interact with each other in highly nontrivial ways. For example, Lamarckian inheritance, an early but now disfavored \autocite{weismann1893germ} theory of evolution, posited that behaviors learned by an individual within its lifetime could be directly transmitted to its progeny so that they would be available as instincts soon after birth. We now know however that known heritable characters are primarily transmitted to the next generation through the genotype. However, over a century ago, Baldwin \autocite{mark1896new} conjectured an alternate mechanism whereby behaviors that are initially learned over a lifetime in early generations of evolution will gradually become instinctual and potentially even genetically transmitted in later generations.  This Baldwin effect seems on the surface like Lamarckian inheritance, but is strictly Darwinian in origin. A key idea underlying this conjecture \autocite{turney2002myths, mayley1996landscapes} is that learning itself comes with likely costs in terms of the energy and time required to acquire skills. 
For example, an animal that cannot learn to walk early in life may be more likely to die, thereby yielding a direct selection pressure on genotypic modifications that can speed up learning of locomotion. More generally, in any environment containing a set of challenges that are fixed over evolutionary timescales, but that also come with a fitness cost for the duration of learning within a lifetime, evolution may find genotypic modifications that lead to faster phenotypic learning. Previous simulations of learning and evolution have provided toy instantiations of the Baldwin effect in highly simplified scenarios \autocite{hinton1987learning, ackley1991interactions, anderson1995learning}. However, biologists have long conjectured that the Baldwin effect might hold at the level of morphological evolution and sensorimotor learning in complex environments \autocite{turney2002myths, waddington1942canalization}. But despite the prevalence of this conjecture, to date no prior study has demonstrated the Baldwin effect in morphological evolution either {\it in vivo} or {\it in silico}. 

%%%%%%%%%%%%%%%%%%%%%%%%%%%%%%%%%%%%%%%%%%%%%%%%%%%%%%%%%%%%%%%%%%%%%%%%%%%%%%%%%%%%%%%%%%%%%%%%%%%%%%%%%%%%%%%%%%%%%
\begin{figure*}[t]
\centering
\includegraphics[width=0.90\linewidth]{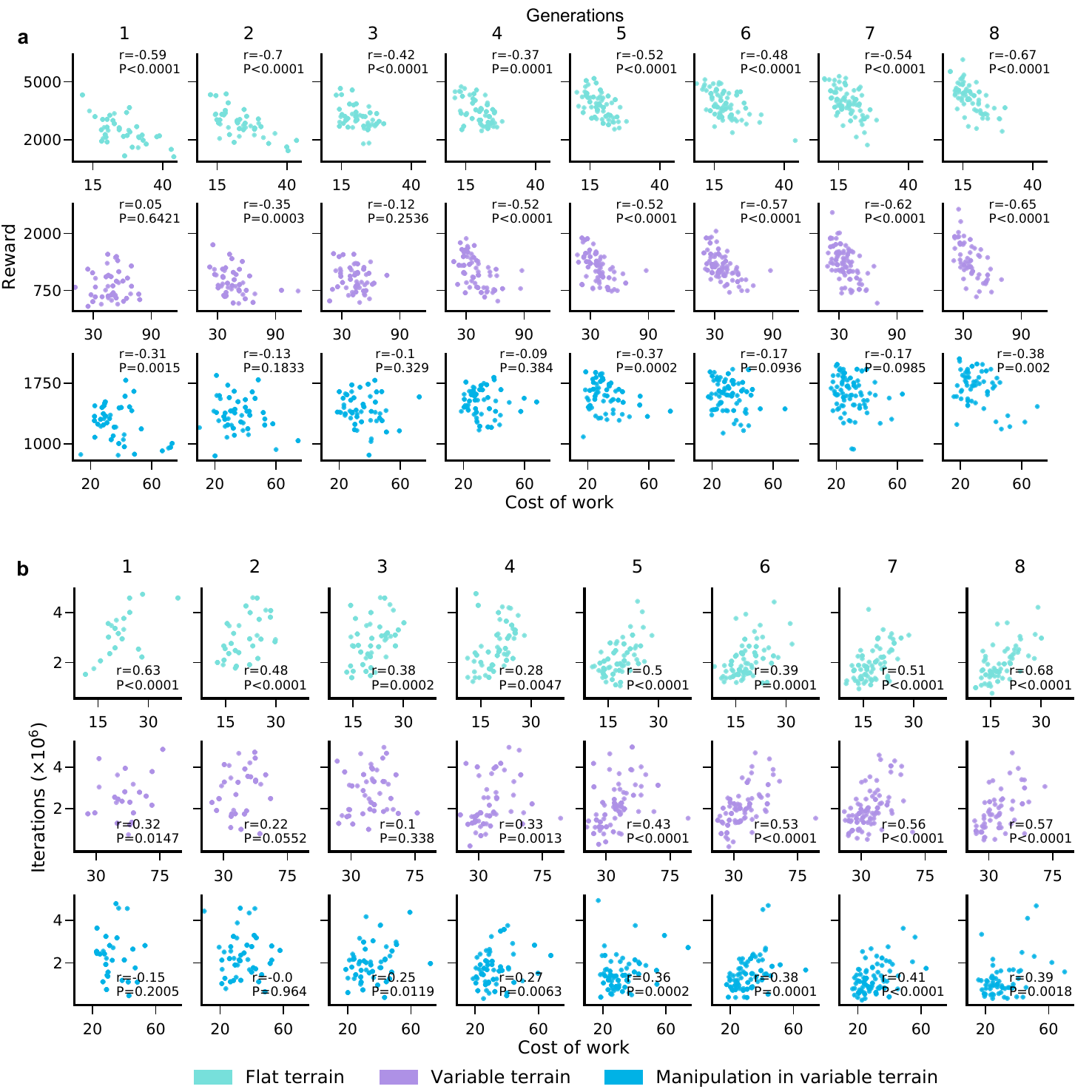}
  \caption{\textbf{Relationship between energy efficiency, fitness and learning speed.} \textbf{a,} Correlation between fitness (reward at the end of lifetime learning) and cost of work for the top $100$ agents across $3$ evolutionary runs. \textbf{b,} Correlation between learning speed (iterations required to achieve the $75\textsuperscript{th}$ percentile fitness of the initial population same as Fig~\ref{fig:baldwin}a) and cost of work for same top $100$ agents as in (\textbf{a}). Across all generations, morphologies which are more energy efficient perform better (negative correlation) and learn faster (positive correlation). (\textbf{a}, \textbf{b}) Shown are the correlation coefficients (r) and P values obtained from two-tailed Pearson’s correlation.}
\label{fig:energy_mech}
\vspace{-3mm}
\end{figure*}
%%%%%%%%%%%%%%%%%%%%%%%%%%%%%%%%%%%%%%%%%%%%%%%%%%%%%%%%%%%%%%%%%%%%%%%%%%%%%%%%%%%%%%%%%%%%%%%%%%%%%%%%%%%%%%%%%%%%%

In our simulations, we find the first evidence for the existence of a morphological Baldwin effect, as reflected by a rapid reduction over generations in the learning time required to achieve a criterion level of fitness for the top $100$ agents in all three environments (Fig.~\ref{fig:baldwin}a). Remarkably, within only 10 generations, average learning time is cut in half. As an illustrative example of how learning is accelerated, we show the learning curves for different generations of an agent evolved in FT (Fig.~\ref{fig:baldwin}d). The $8\textsuperscript{th}$ generation agent not only outperforms the $1\textsuperscript{st}$ generation agent by a factor of $2$ at the end of learning, but can also achieve the final fitness of the first generation agent in $1/5\textsuperscript{th}$ the time.  Moreover, we note that we do not have any explicit selection pressure in our simulations for fast learning, as the fitness of a morphology is determined solely by its performance at the end of learning. Nevertheless, evolution {\it still} selects for faster learners without any direct selection pressure for doing so. Thus we actually discover a stronger form of the Baldwin effect than has been previously conjectured in the literature \autocite{turney2002myths, mayley1996landscapes} by demonstrating that an explicit selection pressure for the speed of skill acquisition is not necessary for the Baldwin effect to hold.  Intriguingly, the existence of this morphological Baldwin effect could be exploited in future studies to create embodied agents with lower sample complexity and higher generalizability. 

\section*{A mechanistic underpinning for morphological intelligence and the strong Baldwin effect}
We next search for potential mechanistic basis for how evolution may both engender morphological intelligence (Fig.~\ref{fig:eval_suite}b,c) as well select for faster learners without any direct selection pressure for learning speed (i.e. the stronger form of the Baldwin effect in Fig.~\ref{fig:baldwin}a). We hypothesize, along the lines of conjectures in embodied cognition\autocite{pfeifer2001understanding, brooks1991new,bongard2014morphology}, that evolution discovers morphologies that can more efficiently exploit the passive dynamics of physical interactions between the agent body and the environment, thereby simplifying the problem of learning to control, which can both enable better learning in novel environments (morphological intelligence), and faster learning over generations (Baldwin effect). Any such intelligent morphology is likely to exhibit the physical properties of both energy efficiency and passive stability, and so we examine both properties.  

We define energy efficiency as the amount of energy spent per unit mass to accomplish a goal (see Appendix D). Surprisingly, without any direct selection pressure for energy efficiency, evolution nevertheless selected for energy efficient morphological solutions (Fig.~\ref{fig:baldwin}c).  We verify such energy efficiency is not achieved simply by reducing limb densities (Fig.~\ref{fig:morphological_descriptior}e). On the contrary, across all three environments the total body mass actually {\it increases} suggesting that energy efficiency is achieved by selecting for morphologies which more effectively leverage the passive physical dynamics of body-environment interactions. Moreover, morphologies which are more energy efficient perform better (Fig.~\ref{fig:energy_mech}a) and learn faster (Fig.~\ref{fig:energy_mech}b) at any fixed generation. Similarly, evolution selects more passively stable (see Appendix D) morphologies over time in all $3$ environments, though the fraction of stable morphologies is higher in VT/MVT relative to FT, indicating higher relative selection pressure for stability in these more complex environments (Fig.~\ref{fig:baldwin}b). Thus, over evolutionary time, both energy efficiency (Fig.~\ref{fig:baldwin}c) and stability (Fig.~\ref{fig:baldwin}b) improve in a manner that is tightly correlated with learning speed (Fig.~\ref{fig:baldwin}a). 

These correlations suggest that energy efficiency and stability may be key physical principles that partially underpin both the evolution of morphological intelligence and the Baldwin effect. With regards to the Baldwin effect, variations in energy efficiency lead to positive correlations across morphologies between two distinct aspects of learning curves: the performance at the end of a lifetime, and the speed of learning at the beginning. Thus evolutionary processes that {\it only} select for the former will implicitly {\it also} select for the latter, thereby explaining the stronger form of the evolutionary Baldwin effect that we observe. With regards to morphological intelligence, we note that MVT and VT agents possess more intelligent morphologies compared to FT agents as evidenced by better performance in test tasks (Fig.~\ref{fig:eval_suite}b), especially with reduced learning iterations (Fig.~\ref{fig:eval_suite}c). Moreover, VT/MVT agents are also more energy efficient compared to FT agents (Fig.~\ref{fig:eval_suite}d). An intuitive explanation of this differential effect of environmental complexity is that the set of subtasks that must be solved accumulates across environments from FT to VT to MVT. Thus, MVT agents must learn to solve more subtasks than FT agents in the same amount of learning iterations. This may result in a higher implicit selection pressure for desirable morphological traits like stability and energy efficiency in MVT/VT agents as compared to FT agents. And in turn these traits may enable better, faster, and more energy efficient performance in novel tasks for MVT/VT agents relative to FT agents (Fig.~\ref{fig:eval_suite}b-d).

\section*{Conclusion}

Thus overall the large-scale simulations made possible by DERL yield scientific insights into how learning, evolution and environmental complexity can interact to generate intelligent morphologies that can simplify control by leveraging the passive physics of body-environment interactions. Intriguingly, we find that the fitness of an agent can be rapidly transferred within a few generations of evolution from its phenotypic ability to learn to its genotypically encoded morphology through a Baldwin effect. These evolved morphologies in turn endow agents with better and faster learning capacities in many novel tasks through embodied morphological intelligence, likely realized through increased passive stability and energy efficency. Indeed this Baldwinian transfer of intelligence from phenotype to genotype has been conjectured to free up phenotypic learning resources to learn more complex behaviors in animals \autocite{waddington1942canalization}, including the emergence of language\autocite{deacon1998symbolic} and imitation \autocite{giudice2009programmed} in humans. This suggests that just as our large scale simulations of learning and evolution can speed up reinforcement learning through the emergence of morphological intelligence, further large-scale explorations of learning and evolution in other contexts may yield new scientific insights into the emergence of rapidly learnable intelligent behaviors, as well as new engineering advances in our ability to instantiate them in machines.
%%%%%%%%%%%%%%%%%%%%%%%%%%%%%%%%%%%%%%%%%%%%%%%%%%%%%%%%%%%%%%%%%%%%%%%%%%%%%%%%%%%%%%%%%%%%%%%%%%%%%%%%%%%%%%%%%%%%%
%%% Main Paper Ends
%%%%%%%%%%%%%%%%%%%%%%%%%%%%%%%%%%%%%%%%%%%%%%%%%%%%%%%%%%%%%%%%%%%%%%%%%%%%%%%%%%%%%%%%%%%%%%%%%%%%%%%%%%%%%%%%%%%%%

%%%%%%%%%%%%%%%%%%%%%%%%%%%%%%%%%%%%%%%%%%%%%%%%%%%%%%%%%%%%%%%%%%%%%%%%%%%%%%%%%%%%%%%%%%%%%%%%%%%%%%%%%%%%%%%%%%%%%
%%% Appendix
%%%%%%%%%%%%%%%%%%%%%%%%%%%%%%%%%%%%%%%%%%%%%%%%%%%%%%%%%%%%%%%%%%%%%%%%%%%%%%%%%%%%%%%%%%%%%%%%%%%%%%%%%%%%%%%%%%%%%
\newpage
%%%%%%%%%%%%%%%%%%%%%%%%%%%%%%%%%%%%%%%%%%%%%%%%%%%%%%%%%%%%%%%%%%%%%%%%%%%%%%%%%%%%%%%%%%%%%%%%%%%%%%%%%%%%%%%%%%%%%
%%% Appendix figures
%%%%%%%%%%%%%%%%%%%%%%%%%%%%%%%%%%%%%%%%%%%%%%%%%%%%%%%%%%%%%%%%%%%%%%%%%%%%%%%%%%%%%%%%%%%%%%%%%%%%%%%%%%%%%%%%%%%%%

\clearpage

\clearpage
%%%%%%%%%%%%%%%%%%%%%%%%%%%%%%%%%%%%%%%%%%%%%%%%%%%%%%%%%%%%%%%%%%%%%%%%%%%%%%%%%%%%%%%%%%%%%%%%%%%%%%%%%%%%%%%%%%%%%
%%% Appendix figures
%%%%%%%%%%%%%%%%%%%%%%%%%%%%%%%%%%%%%%%%%%%%%%%%%%%%%%%%%%%%%%%%%%%%%%%%%%%%%%%%%%%%%%%%%%%%%%%%%%%%%%%%%%%%%%%%%%%%%

%%%%%%%%%%%%%%%%%%%%%%%%%%%%%%%%%%%%%%%%%%%%%%%%%%%%%%%%%%%%%%%%%%%%%%%%%%%%%%%%%%%%%%%%%%%%%%%%%%%%%%%%%%%%%%%%%%%%%
%%% Bibliography
%%%%%%%%%%%%%%%%%%%%%%%%%%%%%%%%%%%%%%%%%%%%%%%%%%%%%%%%%%%%%%%%%%%%%%%%%%%%%%%%%%%%%%%%%%%%%%%%%%%%%%%%%%%%%%%%%%%%%
\printbibliography
\clearpage
%%%%%%%%%%%%%%%%%%%%%%%%%%%%%%%%%%%%%%%%%%%%%%%%%%%%%%%%%%%%%%%%%%%%%%%%%%%%%%%%%%%%%%%%%%%%%%%%%%%%%%%%%%%%%%%%%%%%%

\section*{Appendix A: Evolutionary setup}
\noindent \textbf{Distributed Asynchronous Evolution.} Simultaneously evolving and learning embodied agents with many degrees of freedom  that can perform complex tasks, using only low-level egocentric sensory inputs, required developing a highly parallel and efficient computational framework which we call Deep Evolutionary Reinforcement Learning (DERL). Each evolutionary run starts with an initial population of $P$ agents (here $P=576$) with unique randomly generated morphologies (described in more detail below) chosen to encourage diverse solutions, as evidenced in (Fig.~\ref{fig:muller}b), and prevent inefficient allocation of computational resources on similar morphologies. Controllers for all $P$ initial morphologies are learned in parallel for $5$ million agent-environment interactions (learning iterations) each, and the average reward attained over approximately the last $100,000$ iterations at the end of lifetime learning yields a fitness function over morphologies. Starting from this initial population, nested cycles of evolution and learning proceed in an asynchronous parallel fashion.

Each evolutionary step consists of randomly selecting $T=4$ agents from the current population to engage in a tournament \autocite{goldberg1991comparative, real2019regularized}. In each tournament, the agent morphology with the highest fitness among the $4$ is selected to be a parent.  Its morphology is then mutated to create a child which then undergoes lifetime learning to evaluate its fitness. Importantly, the child starts lifetime learning with a randomly initialized controller, so that only morphological information is inherited from the parent. Such {\it tabula rasa} RL can be extremely sample inefficient especially in complex environments.  Hence, we further parallelize experience collection for training each agent over $4$ CPU cores. $288$ such tournaments and child training are run asynchronously and in parallel over $288$ workers each consisting of $4$ CPUs, for a total of $1152$ CPUs.  The entire computation takes place over $16$ Intel Xeon Scalable Processors (Cascade Lake) each of which has $72$ CPUs yielding the total of $1152$ CPUs. To keep the population size $P$ the same, the oldest member of the population is removed after a child is added \autocite{real2019regularized}. Moreover, this design choice also makes the framework highly fault tolerant; if any compute node fails, a new node can be brought online without affecting the current population. In contrast, if population size is maintained by removing least fit individual in a tournament, compute node failures will need additional book keeping to maintain total population size. Fault tolerance significantly reduces the cost to run DERL on cloud services by leveraging spare unused capacity (spot instances) which is often up to $90\%$ cheaper compared to on demand instances.

\noindent \textbf{UNIversal aniMAL (UNIMAL) design space.} The design of any space of morphologies is subject to a stringent tradeoff between the richness and diversity of realizable morphologies and the computational tractability of finding successful morphologies by evaluating it's fitness. Here, we introduce the UNIMAL design space (Fig.~\ref{fig:system}d), which is an efficient search space over morphologies that introduces minimal constraints  while containing physically realistic morphologies and gaits that can learn locomotion and mobile manipulation.
Our genotype is a kinematic tree, or a directed acyclic graph, corresponding to a hierarchy of articulated 3D rigid parts connected via motor actuated hinge joints. Nodes of the kinematic tree consist of two component types: a sphere representing the head which forms the root of the tree, and cylinders representing the limbs of the agent. Evolution proceeds through asexual reproduction via an efficient set of three classes of mutation operations (Fig.~\ref{fig:system}b) that: (1) either shrink or grow the kinematic tree by starting from a sphere and growing or deleting limbs (grow limb(s), delete limb(s)); (2) modify the physical properties of existing limbs, like their lengths and densities (mutate density, limb params); (3) modify the properties of joints between limbs, including degrees of freedom (DoF), angular limits of rotation, and gear ratios (mutate DoF, joint angle, and gear). During the population initialization phase a new morphology is created by first sampling a total number of limbs to grow and then applying mutation operations until the agent has the desired number of limbs. We now provide a description of the mutation operations in detail:

\noindent \textit{Grow limb(s)}: This mutation operation grows the kinematic tree by adding at most $2$ limbs at a time. We maintain a list of locations where a new limb can be attached. 
The list is initialized with center of the root node. To add a new limb we randomly sample an attachment location from a uniform distribution over possible locations, and randomly sample the number of limbs to add as well as the limb parameters. Limb parameters (Table~\ref{table:unimal_hyper}) include radius, height, limb density, and orientation w.r.t. to the parent limb. We enforce that all limbs have the same density, so only the first grow limb mutation samples limb density, and all subsequent limbs have the same density. However, due to the mutate density operation, all limb densities can change simultaneously under this mutation from a parent to a child. We also only allow limb orientations which ensure that the new limb is completely below the parent limb; i.e. the kinematic tree can only grow in the downward direction. The addition of a limb is considered successful, if after attachment of the new limb the center of mass of the agent lies on the saggital plane and there are no self intersections. Self intersections can be determined by running a short simulation and detecting collisions between limbs. If the new limb collides with other limbs at locations other than the attachment site, the mutation is discarded. Finally, if the limb(s) were successfully attached we update the list of valid attachment locations by adding the mid and end points of the new limb(s). Symmetry is further enforced by ensuring that if a pair of limbs were added then all future mutation operations will operate on both the limbs.

\noindent \textit{Delete limb(s)}: This mutation operation only affects leaf nodes of the kinematic tree. Leaf limb(s) or end effectors are randomly selected and removed from the kinematic tree in a manner that ensures symmetry. 

\noindent \textit{Mutate limb parameters}: A limb is modelled as cylinder which is parameterized by it's length and radius (Table~\ref{table:unimal_hyper}). Each mutation operation first selects a limb or pair of symmetric limbs to mutate and then randomly samples new limb parameters for mutation.

\noindent \textit{Mutate density}: In our design space, all limbs have the same density. To mutate the density we randomly select a new density value (Table~\ref{table:unimal_hyper}). Similarly, we can also mutate the density of the head.

\noindent \textit{Mutate DoF, gear and joint angle}: We describe the three mutations affecting the joints between limbs together, due their similarity. Two limbs are connected by motor actuated hinge joints. A joint is parameterized by it's axis of rotation, joint angle limits and motor gear ratio\autocite{todorov2012mujoco}. 
There can be at most two hinge joints between two limbs.  In MuJoCo\autocite{todorov2012mujoco}, each limb can be described in its own frame of reference in which the limb is extended along the $z$-axis.  In this same frame of reference of the child limb, the $2$ possible axes of rotations of the two hinge joints between the child and parent limbs, correspond to the $x$-axis or the $y$-axis. The main thing this precludes is rotations of a limb about its own axis. 

While attaching a new limb all joint parameters are selected from a predetermined list of possible values (Table~\ref{table:unimal_hyper}). 
Each mutation operation first selects a limb or pair of symmetric limbs to mutate and then modifies the corresponding parameter by a randomly chosen value.

%%%%%%%%%%%%%%%%%%%%%%%%%%%%%%%%%%%%%%%%%%%%%%%%%%%%%%%%%%%%%%%%%%%%%%%%%%%%%%%%%%%%%%%%%%%%%%%%%%%%%%%%%%%%%%%%%%%%%
\begin{table}
\begin{center}
\begin{tabular}{lr}
\toprule
\textbf{Hyperparameter} & \textbf{Value}  \\
\midrule
Max limbs & $10$ \\
Limb radius & $0.05$ \\
Limb height & $[0.2, 0.3, 0.4]$ \\
Limb density & $[500, 600, 700, 800, 900, 1000]$ \\
Limb orientation theta & $[0, 45, 90, 135, 180, 225, 270, 315]$ \\
Limb orientation phi & $[90, 135, 180]$ \\
Head radius & $0.10$ \\
Head density & $[500, 600, 700, 800, 900, 1000]$ \\
Joint axis & $[x, y, xy]$ \\
Motor gear range & $[150, 200, 250, 300]$ \\
\multirow{4}{*}{Joint limits} & $[(-30, 0), (0, 30), (-30, 30),$ \\
& $(-45, 45), (-45, 0), (0, 45),$ \\
& $(-60, 0), (0, 60), (-60, 60)$ \\
& $(-90, 0), (0, 90), (-60, 30) (-30, 60)]$ \\
\bottomrule
\end{tabular}
\caption{\textbf{Hyperparameters for UNIMAL design space.} Mutation operations choose a random element from the corresponding list of possible parameters. The set of all possible values of these hyperparameter choices yields an estimate of $10^{18}$ possible morphologies.}
\label{table:unimal_hyper}
\end{center}
\vspace{-8mm}
\end{table}
%%%%%%%%%%%%%%%%%%%%%%%%%%%%%%%%%%%%%%%%%%%%%%%%%%%%%%%%%%%%%%%%%%%%%%%%%%%%%%%%%%%%%%%%%%%%%%%%%%%%%%%%%%%%%%%%%%%%%

\noindent \textbf{Environments.} DERL enables us to simultaneously evolve and learn agents in three environments (Fig.~\ref{fig:system}e) of increasing complexity: (1) Flat terrain (FT); (2) Variable terrain (VT); and (3) Non prehensile manipulation in variable terrain (MVT). We use the MuJoCo \autocite{todorov2012mujoco} physics simulator for all our experiments. We now provide a detailed description of each environment:

\noindent \textit{Flat terrain.} The goal of the agent is to maximize forward displacement over the course of an episode. At the start of an episode an agent is initialized on one end of a square arena of size ($150 \times 150$ square meters ($m^2$)). 

\noindent \textit{Variable terrain.} Similar to FT, the goal of the agent is to maximize forward displacement over the course of an episode. At the start of an episode an agent is initialized on one end of a square arena of size ($100 \times 100 m^2$). In each episodes, a completely new terrain is created by randomly sampling a sequence of obstacles (Fig.~\ref{fig:system}e) and interleaving them with flat terrain. The flat segments in VT are of length $l \in [1, 3] m$ along the desired direction of motion, and obstacle segments are of length $l \in [4, 8] m$.
Each obstacle is created by sampling from a uniform distribution over a predefined range of parameter values. We consider $3$ types of obstacles: 1. Hills: Parameterized by the amplitude $a$ of $\sin$ wave where $a \in [0.6, 1.2]m$. 2. Steps: A sequence of $8$ steps of height $0.2 m$. The length of each step is identical and is equal to one-eight of the total obstacle length. Each step sequence is always 4-steps up followed by 4-steps down. 3. Rubble: A sequence of random bumps created by clipping a repeating triangular sawtooth wave at the top such that the height $h$ of each individual bump clip is randomly chosen from the range $h \in [0.2, 0.3]m$.  Training an agent for locomotion in variable terrain is extremely challenging as prior work \autocite{heess2017emergence} on learning locomotion in a similar terrain for a hand designed $9$ DoF planar 2D walker required $10^7$ agent-environment interactions, despite using curriculum learning and a morphology specific reward function.

\noindent \textit{Manipulation in variable terrain.} This environment is like VT with an arena of size ($60 \times 40 m^2$). However, here the goal of the agent is to move a box (a cube with side length $0.2m$) from it's initial position to a goal location. All parameters for terrain generation are the same as VT. In each episode, in addition to creating a new terrain, both the initial box location and final goal location are also randomly chosen with the constraint that the goal location is always further along the direction of forward motion than the box location. 

These environments are designed to ensure that the number of sub-tasks required to achieve high fitness is higher for more complex environments. Specifically, a FT agent has to only learn locomotion on a flat terrain. In addition, a VT agent needs to also learn to walk on hills, steps and rubble. Finally, along with all the sub-tasks which need to be mastered in VT, a MVT agent should also learn directional locomotion and non prehensile manipulation of objects. The difference in arena sizes is simply chosen to maximize simulation speed while being big enough that agents can't typically complete the task sooner than an episode length of $1000$ agent-environment interactions (iterations). Hence, the arena size for MVT is smaller than VT.

\section*{Appendix B: Learning algorithm}

\textbf{Reinforcement Learning.} The RL paradigm provides a way to learn efficient representations of the environment from high-dimensional sensory inputs, and use these representations to interact with the environment in a meaningful way. At each time step the agent receives an observation $o_t$ that does not fully specify the state ($s_t$) of the environment, takes an action $a_t$, and is given a reward $r_t$. A policy $\pi_\theta(a_t| o_t)$ models the conditional distribution over action $a_t \in A$ given an observation $o_t \in O(s_t)$.
The goal is to find a policy which maximizes the expected cumulative reward $R=\sum_{t=0}^H \gamma^t r_t$ under a discount factor $\gamma\in[0,1)$, where $H$ is the horizon length. 

\noindent\textbf{Observations.} At each time step, the agent senses the world by receiving low level egocentric proprioceptive and exteroceptive observations (Fig.~\ref{fig:system}c). 
Proprioceptive observations depend on the agent morphology and include joint angles, angular velocities, readings of a velocimeter, accelerometer, and a gyroscope positioned at the head, and touch sensors attached to the limbs and head as provided in the MuJoCo\autocite{todorov2012mujoco} simulator. Exteroceptive observations include task specific information like local terrain profile, goal location, and the position of objects and obstacles. 

Information about the terrain is provided as 2D heightmap sampled on a non-uniform grid to reduce the dimensionality of data. The grid is created by decreasing the sampling density as the distance from the root of the body increases\autocite{heess2017emergence}.  All heights are expressed relative to the height of the ground immediately under the root of the agent. 
The sampling points range from $1m$ behind the agent to $4m$ ahead of it along the direction of motion, as well as $4m$ to the left and right (orthogonal to the direction of motion). Note the height map is not provided as input in tasks like patrol, point navigation etc. where the terrain is flat and does not have obstacles. Information about goal location for tasks like point navigation, patrol etc. and the position and velocity of objects like ball/box for manipulation tasks are provided in an egocentric fashion, using the reference frame of the head. 

\noindent \textbf{Rewards.} The performance of RL algorithms is dependent on good reward function design. A common practice is to have certain components of the reward function be morphology dependent \autocite{heess2017emergence}. However, designing morphology dependent reward functions is not feasible when searching over a large morphological design space. 
One way to circumvent this issue is to limit the design space to morphologies with similar topologies\autocite{wang2018neural}. 
But this strategy is ill-suited as our goal is to have an extremely expressive morphological design space with minimal priors and restrictions. 
Hence, we keep the reward design simple, offloading the burden of learning the task from engineering reward design to agent morphological evolution. 

\noindent For FT and VT at each time step $t$ the agent receives a reward $r_t$ given by,
\[ r_t = w_x v_x - w_c \Vert a \Vert^2 \]
where $v_x$ is the component of velocity in the $+x$ direction (the desired direction of motion), $a$ is the input to the actuators, and $w_x$ and $w_c$ weight the relative importance of the two reward components. Specifically, $w_x = 1$, and $w_c = 0.001$. This reward encourages the agent to make forward progress, with an extremely weak penalty for very large joint torques. 
Note that for selecting tournament winners in the evolutionary process, we only compare the forward progress component of the reward i.e. $w_x v_x$. Hence, there is \textbf{no} explicit selection pressure to minimize energy. We adopt similar a strategy for tournament winner selection in MVT.

\noindent  For MVT at each time step $t$ the agent receives a reward $r_t$ given by,
\[ r_t = w_{ao} d_{ao} + w_{og} d_{og} - w_c \Vert a \Vert^2 \]
Here $d_{ao}$ is geodesic distance between the agent and the object (box) in the previous time step minus this same quantity in the current time step.  This reward component encourages the agent to come close to the box and remain close to it. $d_{og}$ is geodesic distance between the object and the goal in previous time step minus this same quantity in the current time step.  This encourages the agent to manipulate the object towards the goal. The final component involving $\Vert a \Vert^2$ provides a weak penalty on large joint torques as before. The weights $w_{ao}$, $w_{og}$, $w_c$ determine the relative importance of the three components. Specifically, $w_{ao} = w_{og} = 100$, and $w_c = 0.001$. In addition, the agent is provided a sparse reward of $10$ when it's within $0.75 m$ of the initial object location, and again when the object is within $0.75 m$ of goal location. This sparse reward further encourages the agent to minimize the distance between the object and the goal location while being close to the object.

In addition, we use early termination across all environments when we detect a fall. We consider an agent to have fallen if the head of the agents falls below $50\%$ of it's original height. We found employing this early termination criterion was essential in ensuring diverse gaits. Without early termination, almost all agents would immediately fall and move in a snake like gait.

\noindent\textbf{Policy Architecture.} The agent chooses it's action via a stochastic policy $\pi_\theta$ where $\theta$ are the parameters of a pair of deep neural networks: a policy network which produces an action distribution (Fig.~\ref{fig:system}c), and a critic network which predicts discounted future returns. Each type of observation is encoded via a two layer MLP with hidden dimensions $[64, 64]$. The encoded observations across all types are then concatenated and further encoded into a $64$ dimensional vector,  which is finally passed into a linear layer to generate the parameters of a Gaussian action policy for the policy network and discounted future returns for the critic network. The size of the output layer for the policy network depends on the number of actuated joints. We use $\tanh$ non-linearities everywhere, except for the output layers. The parameters of the networks are optimized using Proximal Policy Optimization\autocite{schulman2017proximal} (PPO). 

%%%%%%%%%%%%%%%%%%%%%%%%%%%%%%%%%%%%%%%%%%%%%%%%%%%%%%%%%%%%%%%%%%%%%%%%%%%%%%%%%%%%%%%%%%%%%%%%%%%%%%%%%%%%%%%%%%%%%
\begin{table}
\begin{center}
\begin{tabular}{lr}
\toprule
\textbf{Hyperparameter} & \textbf{Value}  \\
\midrule
Discount $\gamma$ & $.99$ \\
GAE parameter $\lambda$ & $0.95$ \\
PPO clipping parameter $\epsilon$ & $0.2$ \\
Policy epochs & $4$ \\
Batch size & $512$ \\
Entropy coefficient & $0.01$ \\
Reward normalization & Yes \\
Reward clipping & $[-10, 10]$ \\
Observation normalization & Yes \\
Observation clipping & $[-10, 10]$ \\
Timesteps per rollout & $128$\\
\# Workers & $4$\\
\# Environments & $32$\\
Total timesteps & $5 \times 10^6$\\
Optimizer    & Adam  \\ 
Initial learning rate & $0.0003$ \\
Learning rate schedule & Linear decay \\
Gradient clipping ($l_2$ norm) & $0.5$ \\
Clipped value function & Yes \\
Value loss coefficient & $0.5$ \\
\bottomrule
\end{tabular}
\caption{\textbf{PPO hyperparameters.}}
\label{table:ppo_hyper}
\end{center}
\vspace{-8mm}
\end{table}
%%%%%%%%%%%%%%%%%%%%%%%%%%%%%%%%%%%%%%%%%%%%%%%%%%%%%%%%%%%%%%%%%%%%%%%%%%%%%%%%%%%%%%%%%%%%%%%%%%%%%%%%%%%%%%%%%%%%%

\noindent\textbf{Optimization.} Policy gradient methods are a popular class of algorithms for finding the policy parameters $\theta$ which maximize $R$ via gradient ascent. 
Vanilla policy gradient \autocite{williams1992simple} (VPG) is given by $L = \mathbb{E}[\hat{A}_t \nabla_\theta \log \pi_\theta]$, where $\hat{A}_t$ is an estimate of the advantage function. 
VPG estimates can have high variance and be sensitive to hyperparameter changes. 
To overcome this PPO \autocite{schulman2017proximal} optimizes a modified objective $L = \mathbb{E}\left[\min(l_t(\theta)\hat{A}_t, \textrm{clip}(l_t(\theta), 1 - \epsilon, 1 + \epsilon)\hat{A}_t\right]$, where $l_t(\theta) = \frac{\pi_\theta(a_t|o_t)}{\pi_{old}(a_t|o_t)}$ denotes the likelihood ratio between new and old policies used for experience collection.
We use Generalized Advantage Estimation \autocite{Schulmanetal_ICLR2016} to estimate the advantage function. 
The modified objective keeps $l_t(\theta)$ within $\epsilon$ and functions as an approximate trust-region optimization method; allowing for the multiple gradient updates for a mini-batch of experience, thereby preventing training instabilities and improving sample efficiency.  
We adapt an open source \autocite{pytorchrl} implementation of PPO (see Table~\ref{table:ppo_hyper} for hyperparamter values). 
We keep the number of learning iterations the same across all evolutionary environments. In all environments, agents have $5$ millions learning iterations to perform lifetime learning. 

\section*{Appendix C: Evaluation task suite}

A key contribution of our work is to take a step towards quantifying morphological intelligence. Concretely, we compute how much a morphology facilitates the process of learning a large set of test tasks. We create a suite of $8$ tasks (Fig.~\ref{fig:eval_suite}a) categorized into $3$ domains testing agility (patrol, point navigation, obstacle and exploration), stability (escape and incline) and manipulation (push box incline and manipulate ball) abilities of the agent morphologies. Controllers for each task are learned from scratch, thus ensuring that differences in performance are solely due to differences in morphologies.
Note that form RL perspective both task and environment are the same i.e. both are essentially markov decision processes. Here, we use the term environment to distinguish between evolutionary task and test tasks. We now provide a description of the evaluation tasks and rewards used to train the agent. 

\noindent \textbf{Patrol:} The agent is tasked with running back and forth between two goal locations $10m$ apart along the $x$ axis. Success in this task requires the ability to move fast for a short duration and then quickly change direction repeatedly. At each time step the agent receives a reward $r_t$ given by,
\[ r_t = w_{ag} d_{ag} - w_c \Vert a \Vert^2 \]
where $d_{ag}$ is geodesic distance between the agent and the goal in the previous time step minus this same quantity in the current time step, $w_{ag} = 100$, and $w_c = 0.001$. In addition, when the agent is within $0.5m$ of the goal location, we flip the goal location and provide the agent a sparse reward of $10$. 

\noindent \textbf{Point Navigation:} An agent is spawned at the center of a flat arena ($100\times100$ $m^2$). In each episode, the agent has to reach a random goal location in the arena. Success in this task requires the ability to move in any direction. The reward function is similar to the patrol task.

\noindent \textbf{Obstacle:} The agent has to traverse a dense area of static obstacles and reach the end of the arena. 
The base and height of each obstacle varies between $0.5 m$ to $3 m$. The environment is a rectangular flat arena ($150 \times 60$ $m^2$) with $50$ random obstacles initialized at the start of each episode. Success in this task requires the ability to quickly maneuver around obstacles. The obstacle information is provided in the form of terrain height map. The reward function is similar to that of locomotion in FT.

\noindent \textbf{Exploration:} The agent is spawned at the center of a flat arena ($100\times100 m^2$). The arena is discretized into grids of size $(1\times1 m^2)$ and the agent has to maximize the number of distinct squares visited. At each time step agent receives, 
\[ r_t = w_{e} (e_t - e_{t-1})  - w_c \Vert a \Vert^2 \]
where $e_t$ denotes total number of locations explored till time step $t$, $w_{e} = 1$, and $w_c = 0.001$. In addition to testing agility this task is challenging, as unlike in the case of dense locomotion rewards for previous tasks, here the agent gets a sparser reward. 

\noindent \textbf{Escape:} The agent is spawned at the center of a bowl shaped terrain surrounded by small hills\autocite{tassa2020dmcontrol} (bumps). The agent has to maximize the geodesic distance from the start location (escape the hilly region). This task tests the agent's ability to balance itself while going up/down on a random hilly terrain. At each time step the agent receives reward,
\[ r_t = w_{d} d_{as}  - w_c \Vert a \Vert^2 \]
where $d_{as}$ is geodesic distance between the agent and the initial location in the current time step minus this same quantity in the previous time step, $w_{d} = 1$, and $w_c = 0.001$. 

\noindent \textbf{Incline:} The agent is tasked to move on rectangular arena ($150\times40 m^2$) inclined at $10^{\circ}$. Reward is similar to FT.

\noindent \textbf{Push Box Incline:} A mobile manipulation task, where the objective is to push a box (of side length $0.2m$) along an inclined plane. The agent is spawned at the start of a rectangular arena ($80\times40 m$) inclined at $10^{\circ}$. Reward is similar to MVT. 

\noindent \textbf{Manipulate Ball:} A mobile manipulation task, where the objective is to move a ball from a source location to a target location. In each episode a ball (radius $0.2 m$) is placed at a random location on a flat square arena ($30\times30 m$) and the agent is spawned at the center. This task poses a challenging combination of locomotion and object manipulation, since the agent must rely on complex contact dynamics to manipulate the movement of the ball while also maintaining balance. Reward is similar to MVT.

\section*{Appendix D: Evaluation}
\textbf{Reporting Methodology.} The performance of RL algorithms is known to be strongly dependent on the choice of the seed for random number generators \autocite{henderson2018deep}. To control for this variation, within an evolutionary run we use the same seed for all lifetime learning across all morphologies. However we take several steps to ensure robustness to this choice. First, we repeat each evolutionary run $3$ times for each environment with different random seeds. Then to find the best morphologies for each environment in a manner that is robust to choice of seed, we select the top $3$ agents from all surviving lineages across all $3$ evolutionary runs. Typically, in a single evolutionary run we find $15$ to $20$ surviving lineages, yielding a total of $135$ to $180$ good morphologies per environment (at $3$ per lineage over $3$ evolutionary runs). Then we further train these morphologies $5$ times with $5$ entirely new random seeds. This final step ensures robustness to choice of seed without having to run evolution many times. Finally, we select the $100$ best agents in these new training runs for each environment. These $100$ agents are used to generate the data shown in Fig.~\ref{fig:muller}a, Fig.~\ref{fig:baldwin}a, c, and Fig.~\ref{fig:energy_mech}. We also compare the performance of the top $10$ out of these $100$ agents across the suite of $8$ test tasks (Fig.~\ref{fig:eval_suite}b-d).
For all test tasks we use the same network architecture, hyperparameter values and learning procedure as used during evolution, and train the controller from scratch with $5$ random seeds.

\noindent \textbf{Cost of Work.} Cost of transportation\autocite{von1950price} (COT) is a dimensionless measure that quantifies how much energy is applied to a system of a specified mass $M$ in order to move the system a specified distance $D$. That is,
\[ COT = \frac{E}{MgD} \]
where $E$ is the total energy consumption for travelling distance $D$, $M$ is the total mass of the system, and $g$ is the acceleration due to gravity. COT and it's variants have been used in a wide range of domains to compare energy efficient motion of different robotic systems \autocite{Siciliano2016SpringerHO}, vehicles\autocite{von1950price} and animals\autocite{alexander2005models}. We note that COT essentially measures energy spent per unit mass per unit distance, as the normalization factor $g$ required to make this measure dimensionless is the same for all systems. We adapt this metric to more general RL tasks to measure energy per unit mass per unit reward instead of energy per unit mass per unit distance. That is we define a cost of work (COW) by,
\[ COW = \frac{E}{Mgr} \]
where $E$ is the energy spent, $M$ is the mass, and $r$ is the reward achieved. For most locomotion tasks like locomotion in FT/VT, patrol, obstacle, escape and incline where reward is proportional to distance travelled; COW and COT are essentially the same albeit with different units. We measure energy as the absolute sum of all joint torques\autocite{yu2018learning}. This definition was used to compute energy efficiency (with lower COW indicating higher energy efficiency) in both evolutionary environments (Fig.~\ref{fig:baldwin}c) and test tasks (Fig.~\ref{fig:eval_suite}b-d).

\noindent \textbf{Stability.} Informally, passive stability is the ability to stand without falling and is achieved via mechanical design of the agent/robot. Dynamic stability is ability to move without falling over and is achieved via control. Formally an agent is passively stable, when the centre of mass is inside the support polygon and the polygon’s area is greater than zero \autocite{mcghee1968stability}. The support polygon is the convex hull of all of the agent's contact points with the ground. We measure passive stability by checking if the agent falls over without any control. The agent is initialized at the center of arena and we measure the position of the head at the beginning and after $400$ time steps (a full episode is $1000$ time steps). An agent is passively stable if the head position after $400$ time steps is above $50\%$ of original height. Note that we use the violation of this same condition for early termination of the episode (see Rewards). We use this notion of passive stability in Fig.~\ref{fig:baldwin}b.

\noindent \textbf{Beneficial mutations.} We define a mutation to be beneficial if the difference between the child and parent fitness is above a certain threshold. Although any non-zero increase in fitness is a beneficial mutation, small changes in fitness acquired via RL might not be statistically meaningful, especially since during evolution the fitness is calculated using a single seed. Hence, we heuristically set the threshold as a minimum increase in final average reward by $300$ for FT and $100$ for VT and MVT. Roughly these numbers correspond to the $75\textsuperscript{th}$ percentile in the distribution of fitness increases across all mutations in a given environment. We use this definition of beneficial mutations in Fig.~\ref{fig:muller}b.

\section*{Acknowledgements}
We thank Daniel Yamins, Daniel S. Fisher, Benjamin M. Good and Ilija Radosavovic for feedback on the draft. S.G. thanks the James S. McDonnell and Simons foundations, NTT Research, and an NSF CAREER award for funding. A.G. and L.F-F. thank Adobe, Weichai America Corp and Stanfod HAI for funding.
\newpage
%%%%%%%%%%%%%%%%%%%%%%%%%%%%%%%%%%%%%%%%%%%%%%%%%%%%%%%%%%%%%%%%%%%%%%%%%%%%%%%%%%%%%%%%%%%%%%%%%%%%%%%%%%%%%%%%%%%%%
%%% Appendix section ends
%%%%%%%%%%%%%%%%%%%%%%%%%%%%%%%%%%%%%%%%%%%%%%%%%%%%%%%%%%%%%%%%%%%%%%%%%%%%%%%%%%%%%%%%%%%%%%%%%%%%%%%%%%%%%%%%%%%%%
\end{document}